\documentclass[letterpaper, 10pt,conference]{ieeeconf}  

\IEEEoverridecommandlockouts                              

\usepackage{graphics} 
\usepackage{epsfig} 
\usepackage{mathptmx} 
\usepackage{times} 
\usepackage{amsmath} 
\usepackage{amssymb}  
\usepackage{array}
\usepackage{algorithmicx}
\usepackage{algorithm}
\usepackage{algpseudocode}
\usepackage{multirow}
\usepackage{makecell}
\usepackage{diagbox}
\usepackage{hyperref}
\usepackage{etoolbox}
\usepackage{cite}
\makeatletter
\patchcmd{\@makecaption}
  {\scshape}
  {}
  {}
  {}
\makeatletter
\patchcmd{\@makecaption}
  {\\}
  {.\ }
  {}
  {}
\makeatother

\usepackage{booktabs}
\usepackage{array, caption, threeparttable}
\newsavebox{\measurebox}

\usepackage[caption=false,font=normalsize,labelfont=sf,textfont=sf]{subfig}

\title{\LARGE \bf
ANMS: Asynchronous Non-Maximum Suppression in Event Stream
}

\author{Qianang Zhou$^{1,2}$, Junlin Xiong$^{1}$~\IEEEmembership{Member, IEEE}, Youfu Li$^{2}$~\IEEEmembership{Fellow, IEEE}
\thanks{This work was supported in part by the National Natural Science Foundation of China under Grant 61773357. (\emph{corresponding author: Junlin Xiong})}
\thanks{$^{1,2}$Qianang Zhou is with Department of Automation, University of Science and technology of China, Hefei 230026, China and Department of Mechanical Engineering, City University of Hong Kong, Kowloon, Hongkong 999077, China. {email: zqa1313@mail.ustc.edu.cn and qianazhou2-c@my.cityu.edu.hk}}%
\thanks{$^{1}$Junlin Xiong is with Department of Automation, University of Science and technology of China, Hefei 230026, China. {email: xiong77@ustc.edu.cn}}%
\thanks{$^{2}$Youfu Li is with Department of Mechanical Engineering, City University of Hong Kong, Hong Kong 999077, China. {e-mail: meyfli@cityu.edu.hk}}
}

\begin{document}

\maketitle
\thispagestyle{empty}
\pagestyle{empty}

\begin{abstract}
The non-maximum suppression (NMS) is widely used in frame-based tasks as an essential post-processing algorithm. However, event-based NMS either has high computational complexity or leads to frequent discontinuities. As a result, the performance of event-based corner detectors is limited. This paper proposes a general-purpose asynchronous non-maximum suppression pipeline (ANMS), and applies it to corner event detection. The proposed pipeline extract fine feature stream from the output of original detectors and adapts to the speed of motion. The ANMS runs directly on the asynchronous event stream with extremely low latency, which hardly affects the speed of original detectors. Additionally, we evaluate the DAVIS-based ground-truth labeling method to fill the gap between frame and event.
Evaluation on public dataset indicates that the proposed ANMS pipeline significantly improves the performance of three classical asynchronous detectors with negligible latency. More importantly, the proposed ANMS framework is a natural extension of NMS, which is applicable to other asynchronous scoring tasks for event cameras. The C++ implementation of our ANMS-based detectors will be released here: \href{https://github.com/ZhouQianang/ANMS}{https://github.com/ZhouQianang/ANMS}.

\end{abstract}

\begin{keywords}
Visual Tracking, Computer Vision for Automation, SLAM.
\end{keywords}

\section{Introduction}

Visual Simultaneous Localization and Mapping (VSLAM) refers to simultaneously self-localizing and reconstructing the environment using frame-based cameras. However, frame-based cameras cannot handle fast motion and high dynamic range scenes. The advent of dynamic vision sensors \cite{eventcamera} has opened more opportunities for VSLAM. In contrast to frame-based cameras, event camera pixels respond asynchronously to brightness changes. The new paradigm sensor converts the brightness signal into a stream of events with extremely high temporal resolution. This allows the event camera to deal with fast motion and over/under-exposed scenes. VSLAM systems with the aid of event cameras have shown superior performance \cite{Ultimate}.

Feature-based schemes for pose estimation account for a large proportion of VSLAM systems. For event-based vision, researchers have proposed various asynchronous feature detection algorithms \cite{evFAST,ArcFAST,FAHarris,SITS} to exploit the low-latency advantage of event cameras. Furthermore, several odometry based on asynchronous feature tracking have been proposed recently \cite{GWPHKU:EVIO,PL-EVIO,ESVIO}. Event-based tracker provides high temporal resolution trajectories in challenging scenarios. However, the asynchronous output of the event camera introduces huge challenges to some of the necessary algorithms, such as non-maximum suppression (NMS).

As a crucial post-processing algorithm in frame-based vision, NMS is widely used for tasks such as feature detection, edge detection \cite{canny}, and object detection \cite{ObjectNMS}, etc. The NMS filters out other candidate pixels by keeping only the pixel with the largest response within the local frame window. There is, however, no longer a concept of frame in event streams, and the standard NMS fails to perform directly. In addition, the events collected in the local spatiotemporal window have different timestamps, so that the response of them cannot be appropriately compared. In the absence of the NMS, most detectors filter features only by simple thresholding \cite{evHarris,FAHarris,SITS}. However, applying a uniform threshold to features with different appearances leads to an uneven density of feature streams. As a result, the coarse feature stream considerably affect the tracking accuracy and further affect the localization performance.

\begin{figure}[t]
\centering
\includegraphics[width=0.48\textwidth]{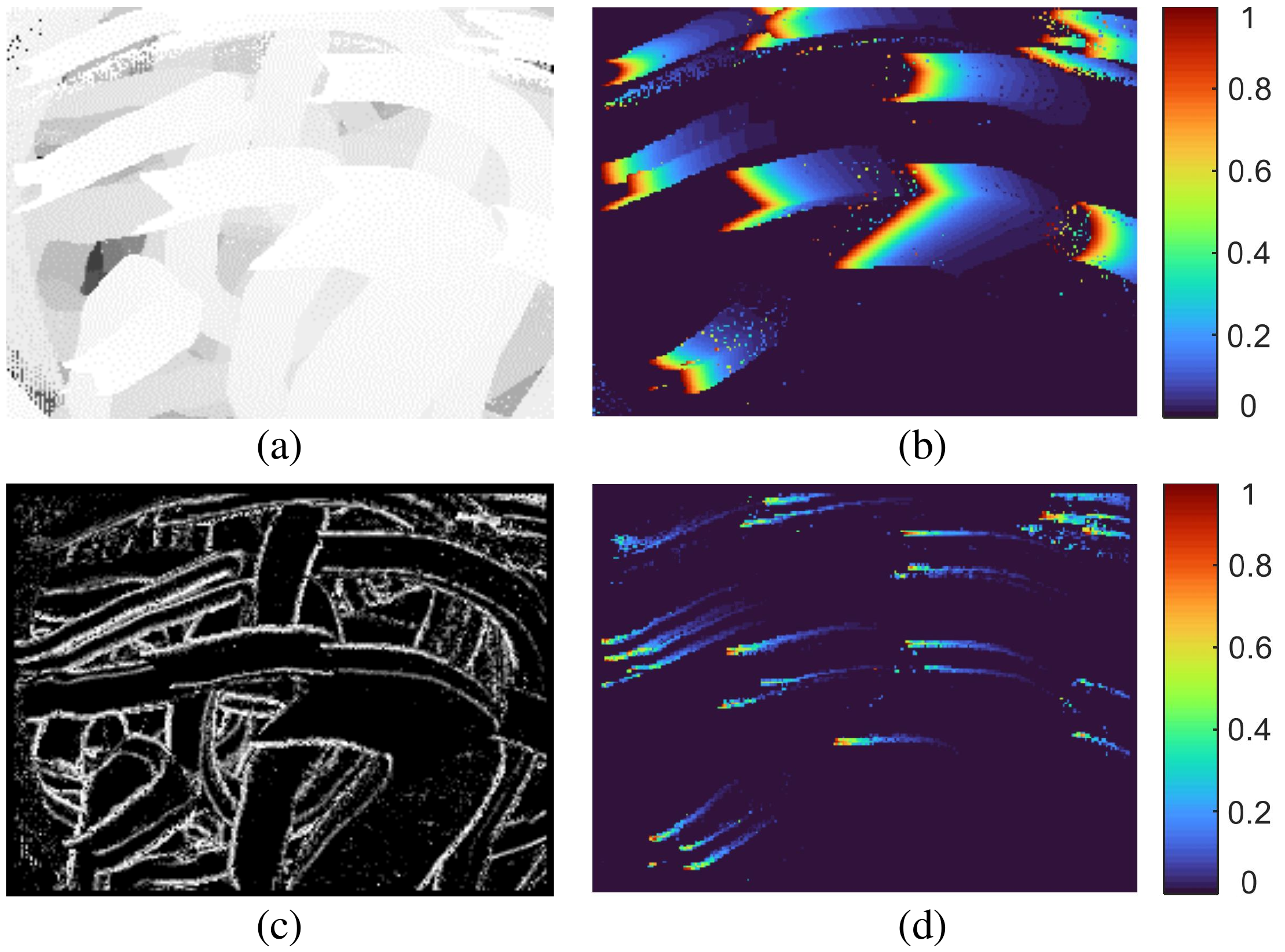}
\caption{Visualizing several essential concepts in the proposed ANMS pipeline at a particular moment (on \textit{shapes\_6dof} dataset). (a) The surface of active events (SAE). (b) The exponential time decay coefficient, calculated according to SAE with $\tau$ equals $0.05s$. (c) The Harris corner score of SAE (SSAE). (d). The exponentially decaying score of SAE (DSSAE).}
\label{fig:SAEs}
\end{figure}

\begin{figure*}[t]
\centering
\includegraphics[width=12cm,height=2.08cm]{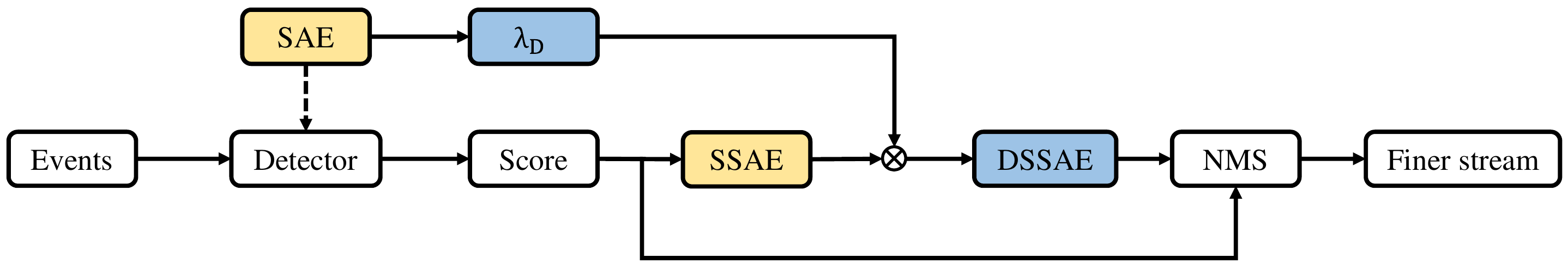}
\caption{The architecture of the proposed ANMS pipeline. The dashed arrow indicates that most asynchronous detectors \cite{evFAST, ArcFAST, FAHarris, SITS} maintain an SAE. The SAE and SSAE (yellow) are maintained in the global image, whereas $\lambda_D$ and DSSAE (blue) are only computed locally.}
\label{fig:ANMSfw}
\end{figure*}

Powered by the need to refine asynchronous output of event algorithms. this paper proposes an asynchronous non-maximum suppression framework (ANMS). ANMS asynchronously maintains two global event states to perform NMS, which suppresses the coarse feature streams to fine. Extensive experiments indicate that the designed framework significantly imporve the performance of original detectors with extremely low latency and memory consumption. More importantly, ANMS is not only suitable for feature detection, but also for other asynchronous scoring tasks in event vision. Specifically, the contributions of this paper are as follows:

\begin{itemize}
\item A general-purpose and adaptive asynchronous non-maximum suppression pipeline, which can be executed swiftly. The evaluation on the public dataset indicates that the ANMS improves the performance of all available detectors.
\item A simple and effective scoring method for event-based FAST algorithms \cite{evFAST, ArcFAST}, inspired by the FAST algorithm \cite{FAST}. The proposed algorithm reuses the intermediate results.
\item A method for evaluating the ground-truth. With the evaluation, we adjust the parameters to imporve the ground-truth.
\end{itemize}

Following related works, we describe the proposed ANMS pipeline in detail. We then demonstrate the ground-truth evaluating and labeling method. Afterwards, we present and analyze the evaluation for ANMS. Finally, we discuss the limitations and future improvements of the method.

\section{Related Work}\label{sec:ReaWo}

This work proposes an ANMS framework with a novel asynchronous event representation, and applies the framework to event-based corner detection. Therefore, this section reviews related works on asynchronous event representation, event-based corner detection, and NMS, respectively.

\textbf{Asynchronous event representation.} There is a general consensus in event-based vision that, individual events carry less information, whereas synchronous representations lose the low-latency advantage of event cameras. As a result, the asynchronous representations with more information are preferred. Surface of Active Events (SAE) \cite{delbruckSAEraw}, also called Time Surface (TS), stores the latest event timestamp for each pixel address, is one of the most important asynchronous representations in event vision. Lagorce et al. \cite{HOTS} derived a more stable representation by exponentially decaying the SAE timestamps, and successfully applied it to several pattern recognition tasks. In order to reduce the sensitivity of SAE to noise, Sironi et al. \cite{HATS} proposed local memory time surface by considering additional event histories. Alzugaray et al. \cite{ACE} proposed sort normalization of SAE to diminish the effect of speed. To speed up the sort normalization SAE, Manderscheid et al. \cite{SITS} proposed a faster representation, Speed-Invariant-Time-Surface (SITS). Baldwin et al. \cite{baldwinTORE} extended SAE to multi-channel representation by storing multiple up-to-date events for each pixel address. Our approach performs ANMS by processing historical events and scores in the form of SAEs.

\textbf{Synchronous corner detection.} Synchronous algorithms integrate events into event frames and apply classical feature detectors. Both Rebecq et al. \cite{RebecqVIO} and Vidal et al. \cite{Ultimate} detected and tracked FAST features \cite{FAST} on the motion-compensated event frames. Alex et al. \cite{ZhuzihaoP} applied the Harris detector on event frames and associated these features between frames. Synchronous detection directly benefits from the frame-based algorithms, but loses the advantage of the event camera. The method presented in this paper is developed for asynchronous detectors.

\textbf{Asynchronous corner detection.} Clady et al. \cite{Clady} search for corners at planar intersections by fitting local spatiotemporal planes. Vasco et al. \cite{evHarris} proposed evHarris, applied the Harris \cite{Harris} detector on the binary SAE. After that, Mueggler et al. \cite{evFAST} proposed evFAST, performed the adapted FAST \cite{FAST} detector on the original SAE. Glover et al. \cite{gloverLUVHARRIS} proposed a faster Harris-based detector. Alzugaray et al. \cite{ArcFAST} accelerated the evFAST and enables it to detect corners greater than 180 degrees, the improved detector is called ArcFAST. Li et al. \cite{FAHarris} combined the ArcFAST with evHarris to trade off speed and accuracy. Manderscheid et al. \cite{SITS} trained a random forest based on SITS to detect stable corners. These works are designed to detect more stable corners. However, due to the absence of NMS, they use a pre-set threshold to filter corners. The proposed ANMS is designed to fill this gap by running directly after asynchronous detectors.

\textbf{Non-maximum suppression.} Scheerlinck et al. \cite{Scheerlinck} established the Harris corner-response states and performed NMS periodically. Yilmaz et al. \cite{Yilmaz} considered that periodically performing NMS extremely penalizes the corner event stream and presents frequent discontinuities. Inspired by, the proposed ANMS framework is executed asynchronously and preserves the continuity of the event stream as much as possible. More importantly, the proposed pipeline is available for other asynchronous scoring tasks.

\section{Method}

\subsection{From NMS to ANMS}\label{sec:ANMSOverview}
Before introducing ANMS, we analyze NMS and give the necessary conditions to extend it to event streams. On the image $I(x,y)$, the corner score of each pixel forms $R(x,y)$. The NMS algorithm classifies pixel $i$ as non-corner unless the score of pixel $i$ is highest within the local window $w_i$. The pixel $i$ meeting Eq. \eqref{con:NMS} has the highest local score.

\begin{equation}
  R(x_i,y_i)=\max\limits_{j\in w_i}\{R(x_j,y_j)\}
  \label{con:NMS}
\end{equation}

Therefore, the first necessary condition for executing asynchronous NMS is corner scores, which is met by most detectors. For detectors without corner scores \cite{evFAST,ArcFAST}, we provide a solution in Section \ref{sec:CSE}, which is one of the contributions of this paper. The second necessary condition is to form corner scores into a comparable frame-like representation. However, all events within the local window have different timestamps and cannot be compared directly. For example, multiple events generated by one corner will suppress each other, resulting in a sharp discontinuity. We illustrate the proposed asynchronous representation Decaying Score of SAE (DSSAE) in Section \ref{sec:DSSAE}, and explain how it works for ANMS in Section \ref{sec:ANMS}.

\subsection{Event Model and Representation}\label{sec:EMR}

The event camera encodes the brightness signal into asynchronous event streams. Specifically, the event camera will output an event as long as any pixel detects a brightness change that exceeds the threshold $C$. For example, the pixel at $(x_i,y_i)$ detects a brightness change at time $t_i$, which will generate an event:
\begin{equation}
  \mathbf{e_i} = (t_i,x_i,y_i,p_i),
\end{equation}
where $p_i \in \{-1,+1\}$ represents the direction of brightness change, $-1$ denotes a decrease, and $+1$ denotes an increase. 

The independent events are usually integrated to an asynchronous representation. As described in Section \ref{sec:ReaWo}, our algorithm is inspired by SAE. The SAE is defined as
\begin{equation}
  SAE:(x,y,p)\mapsto t,
\end{equation}
where $t$ is the timestamp of the latest event with polarity $p$ occurred at $(x, y)$. The SAE is updated asynchronously. For each event, only the pixel corresponding to the event is updated. Finally, note that we maintain SAE according to polarity separately, as in previous works. 

\subsection{Exponentially Decaying Score of SAE}\label{sec:DSSAE}

Events collected in the local window have different timestamps, and their scores cannot be directly compared. Older corner events will suppress newer corner events. Intuitively, the older the event is, the less effect it has on the current decision. Therefore, it is relatively fair to weight the scores with the timestamp information. The final representation is obtained in two steps, (i) collecting the event timestamps and scores, (ii) weighting scores with the timestamps.

We collect the timestamps and scores of events in the same way as SAE. For simplicity, we define `active events' as the set of the latest events at each pixel. The timestamps of active events are stored in the SAE. The scores of active events are stored in a representation which we call Score of SAE (SSAE). Similarly, the SSAE is defined as
\begin{equation}
  SSAE:(x,y,p)\mapsto s,
\end{equation}
where $s$ is the score of the latest event with polarity $p$ occurred at $(x, y)$. In Fig. \ref{fig:SAEs}, we provide snapshots of SAE and SSAE. The SAE shows the uncovered historical motion of scenes. The SSAE highlights the historical motion of corner points. 

\begin{figure}[t]
  \includegraphics[width=0.45\textwidth]{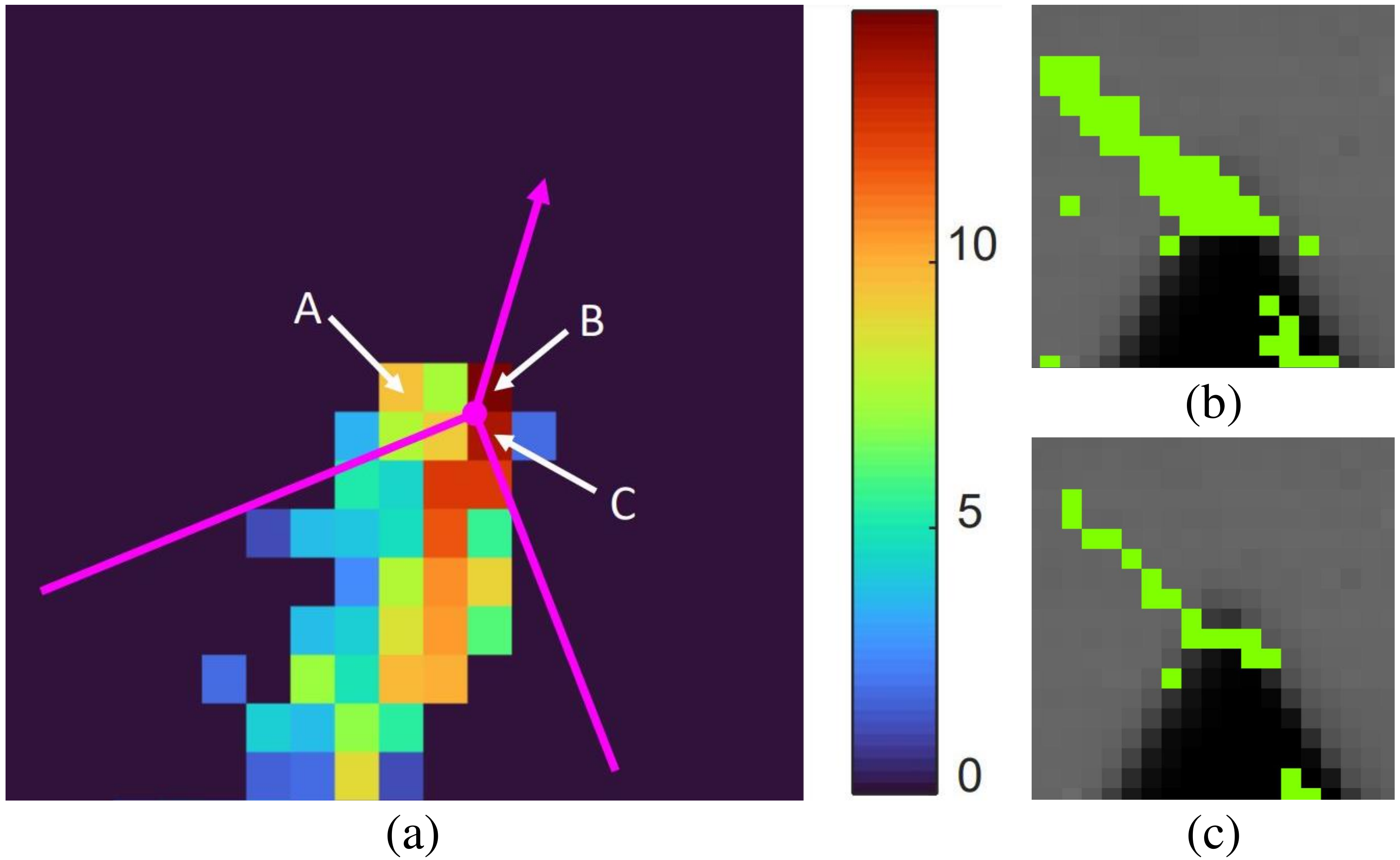}
  \caption{Illustration of performing ANMS at a corner. (a) The DSSAE formed by corner motion. The magenta lines represent the corner shape and the direction of movement. Points A, B, and C represent a new non-corner event, a new corner event, and an old corner event. The corner event stream output by detectors without ANMS (b) and with ANMS (c) in the same corner.}
  \label{fig::ANMSABC}
\end{figure}

\begin{equation}
  \lambda_D(x,y,p)=\exp(-\frac{t-SAE(x,y,p)}{k\tau})
  \label{con:lambda}
  \end{equation}

As analyzed before, putting all events directly together without distinguishing timestamps leads to confusion.Inspired by \cite{HOTS}, we use the exponential kernel to highlight new events. The decay coefficient $\lambda_D$ derived from SAE is defined as Eq. \eqref{con:lambda}, where $t$ is the current time and $k\tau$ is the time constant. We show the $\lambda_D$ in Fig. \ref{fig:SAEs}(b) with new events in red and old events in blue. Compared to the original SAE, $\lambda_D$ increases the difference between old and new events, which allows a clear view of the recent movement.

\begin{equation}
  DSSAE(x,y,p)=\lambda_D(x,y,p)\cdot SSAE(x,y,p)
  \label{con:DSSAE}
\end{equation}

Finally, we use $\lambda_D$ as the weight to obtain the Exponentially Decaying Score of SAE (DSSAE) for ANMS. As defined in Eq. \eqref{con:DSSAE}, the DSSAE is equal to the Hadamard product of $\lambda_D$ and SSAE. The snapshot of DSSAE in Fig. \ref{fig:SAEs}(d) clearly shows the trajectory of corners, and the newer the event, the higher the value is. And for ideal scores, the closer to the center of the trajectory, the higher the DSSAE value is. The two attributes of DSSAE are key to applying it to ANMS.

\subsection{Non-Maximum Suppression by DSSAE}\label{sec:ANMS}

This section describes the process and principle of ANMS. In general, we maintain SAE and SSAE asynchronously and globally, and compute DSSAE within the local window centered on the current event. For an incoming event $\mathbf{e_i} = (t_i,x_i,y_i,p_i)$, the ANMS filters out $\mathbf{e_i}$ if $\mathbf{e_i}$ does not meet Eq. \eqref{con:ANMS}. The architecture of the ANMS pipeline is shown in Fig. \ref{fig:ANMSfw}. 

\begin{equation}
  DSSAE(x_i,y_i,p_i)=\max\limits_{j\in w_i,~p_i=p_j}\{DSSAE(x_j,y_j,p_j)\}
  \label{con:ANMS}
\end{equation}

In the following, we explain the principle of DSSAE for ANMS. Fig. \ref{fig::ANMSABC} illustrates three representative events on DSSAE around a corner, and the colors represent the DSSAE values. The order of their timestamps is $t_C<t_A\le t_B = t$, and the order of original scores is $s_C\approx s_B> s_A>T_d$. We focus on the suppression of events $A$ and $B$ by older event $C$. (i) Event $A$ is non-corner but has a sufficiently high score. The original detector considers it as a corner event because its score exceeds the threshold $T_d$. In ANMS pipeline, the previous corner event $C$ on DSSAE would suppress $A$, because $C$ has a higher score than $A$ locally. (ii) When the corner event $B$ occurs, $B$ has the highest score locally. Although $C$'s original score is as high as $B$'s, $C$'s DSSAE score is exponentially decayed. Therefore, the event $A$ is filtered out and the event $B$ is retained by the pipeline. Afterward, we show the result of ANMS filtering coarse corner stream with an example in Fig. \ref{fig::ANMSABC} (b) and (c). The result indicates that the ANMS tends to retain the center of the trajectory. 

During the processing, the decay coefficient $\lambda_D$ is vital. According to Eq. \eqref{con:lambda}, fast motion leads to smaller timestamp intervals and $\lambda_D$ converges to $1$, and vice versa. Therefore, the $\tau$ is designed to be adaptive to speed to keep a consistent suppression intensity. Previous methods used the inverse of optical flow to estimate $\tau$ \cite{ZhuzihaoP}, which is unsuitable for our task due to the high computation cost. As a simple and effective alternative, we set $\tau$ to the average timestamp interval between the current event and the latest $5$ events within the window. The $k$ is set as $20$ experimentally. 

\subsection{Corner Score of Event}\label{sec:CSE}


We selected three detectors for the evaluation in Sec. \ref{sec:ExpEva}, which are evHarris \cite{evHarris}, evFAST \cite{evFAST}, and ArcFAST \cite{ArcFAST}. 
Neither evFAST nor ArcFAST scored the event, thus we design a scoring method for them in this section.

\begin{figure}[t]
\centering
\includegraphics[width=0.48\textwidth]{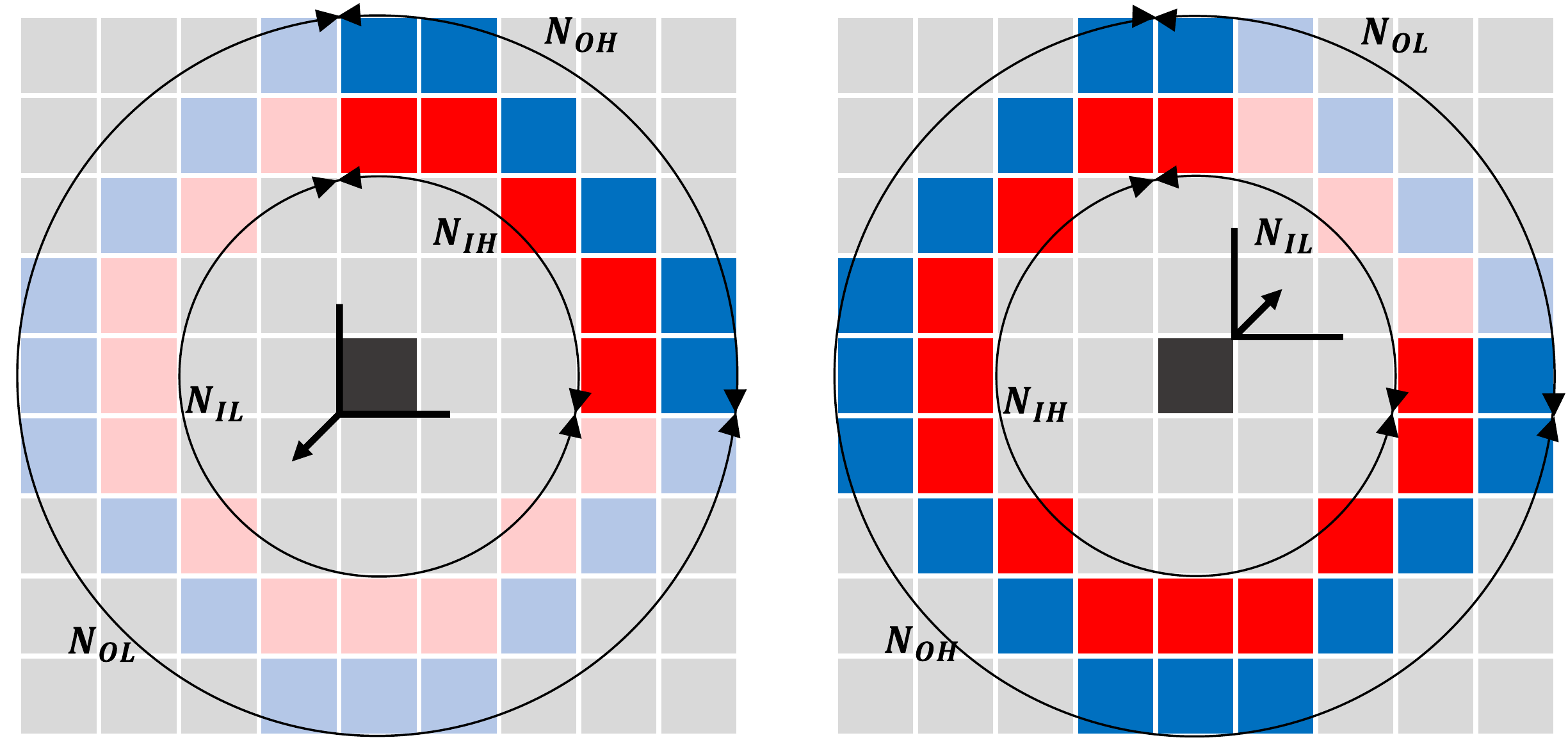}
\caption{The segmentation results of the event-based FAST algorithms on SAE around two corners ($90^\circ$ in left and $270^\circ$ in right), with the arrows indicate the direction of their movement. The inner (red) and outer (blue) circles are divided into higher (brighter color) and lower (lighter color) timestamps segments, respectively. Variables ($N_{IH},N_{IL},N_{OH},N_{OL}$) represents the length of segments. The gray pixels are not involved in the detection of the current pixel (black).}
\label{fig:FASTscore}
\end{figure}

Both of them are adapted from the FAST. The FAST algorithm considers a circle around the candidate pixel and searches for contiguous segment that are brighter/darker than the candidate pixel. Similar to the FAST, they consider circles around the current event on the SAE, as shown in Fig. \ref{fig:FASTscore}. However, the current event (indicated by black pixels) has the highest timestamp on SAE and the comparison with it is non-informative. Therefore, they search for contiguous segment with higher timestamps than the other segments (segments with brighter color in Fig. \ref{fig:FASTscore}). To reduce the sensitivity to noise, the search is performed independently on two concentric circles (the red inner circle and the blue outer circle). The authors consider the current event as a corner if the segments on both circles are within the preset length. To detect corners greater than 180 degrees, ArcFAST set the length to $[3,6]\bigcup [10,13]$ pixels for the inner circle, and $[4,8]\bigcup [13,16]$ pixels for the outer circle.

Fig. \ref{fig:FASTscore} shows the segmentation results of the algorithm on SAE for two corners. The left and right figures represent corners less than and greater than 180 degrees, respectively. Whether on the left or right, the algorithms divide both concentric circles into two segments (with brighter or lighter timestamps). Inspired by FAST \cite{FAST}, we define the corner score of the central event as the sum of the lengths of the two major arcs. Therefore, the score of event-based FAST corner is described as
\begin{equation}
V_{event}=\max\{N_{IH},N_{IL}\}+\max\{N_{OH},N_{OL}\},
\label{con:EFASTSC}
\end{equation}
where the $N_{IH}$/$N_{IL}$ represents the length of the segment with higher/lower timestamps on the inner circle, and the $N_{OH}$/$N_{OL}$ represents the length of the segment with higher/lower timestamps on outer circle.

In terms of computational cost, the scoring method reuses the intermediate result of detector and performs only simple operations. In terms of the ability to describe corners, the closer the central pixel is to the corner, the longer the length of the major arc, the higher the score. Besides, inspired by \cite{ArcFAST}, the score is composed of longer segments instead of segments with higher timestamps, so it applies to less than and greater than 180 degrees corners, like ArcFAST. Take corners shown in Fig. \ref{fig:FASTscore} as examples. For the left figure, the angle is less than 180 degrees, and the score equals the sum of $N_{IL}$ and $N_{OL}$. For the right figure, the angle is greater than 180 degrees, and the score equals the sum of $N_{IH}$ and $N_{OH}$.

\section{Ground-Truth Collection}\label{sec:groundtruth}

The high temporal resolution of the event stream poses a great challenge for manual corner ground-truth labeling. For the DAVIS camera, a frequently-used ground-truth labeling method is described and used in \cite{ArcFAST,FAHarris,Yilmaz}. Likewise, Manderscheid et al. \cite{SITS} designed an automatic corner labeling method for the ATIS camera. We focus on the DAVIS-based method and illustrate our ground-truth method with the DAVIS 240C dataset \cite{dataset} as an example. In this section, we improve the effectiveness of the previous method by evaluating the ground-truth and tuning the parameters.

The method detects Harris corners on the frames of DAVIS and tracks them by KLT \cite{KLT} to obtain feature trajectories. An event is labeled as a corner if its distance from the intensity trajectory is smaller than the threshold. This method uses only frame data to label events, and its effectiveness is debatable for two reasons. First, the event data and frame data have different interpretations of corner \cite{evFAST}, thus the event and intensity corners do not coincide exactly. Most of the previous works use a large distance threshold ($3.5$ pixels) to weaken the discrepancy an thus generate coarse ground-truth.
Second, there is a lack of methods for evaluating ground-truth. As a result, it is not feasible to bring the intensity corner close to the event corner by adjusting the parameters of the labeling algorithm.
 
\begin{figure}[t]
    \centering
    \subfloat[][]{\includegraphics[width=0.23\textwidth]{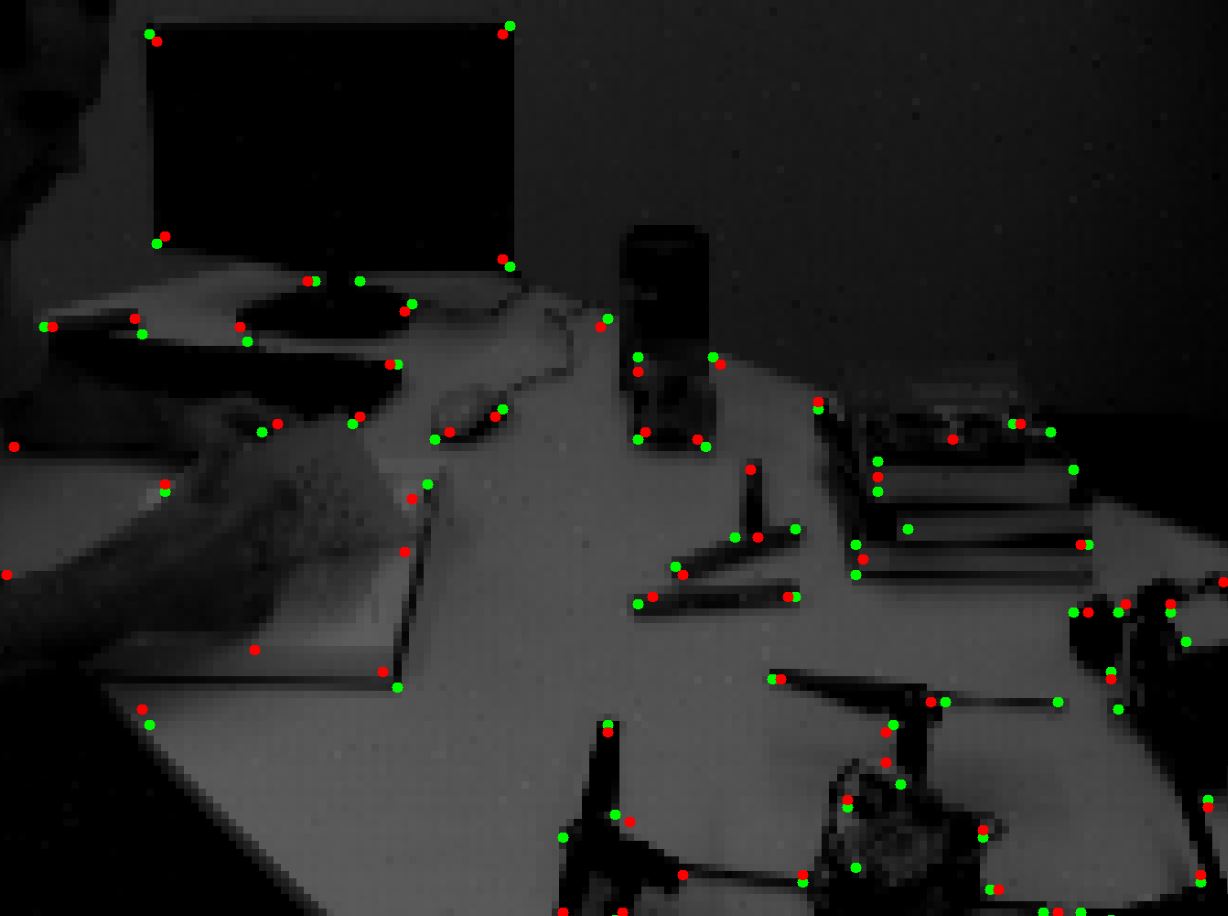}\label{dist}}~
    \subfloat[][]{\includegraphics[width=0.23\textwidth]{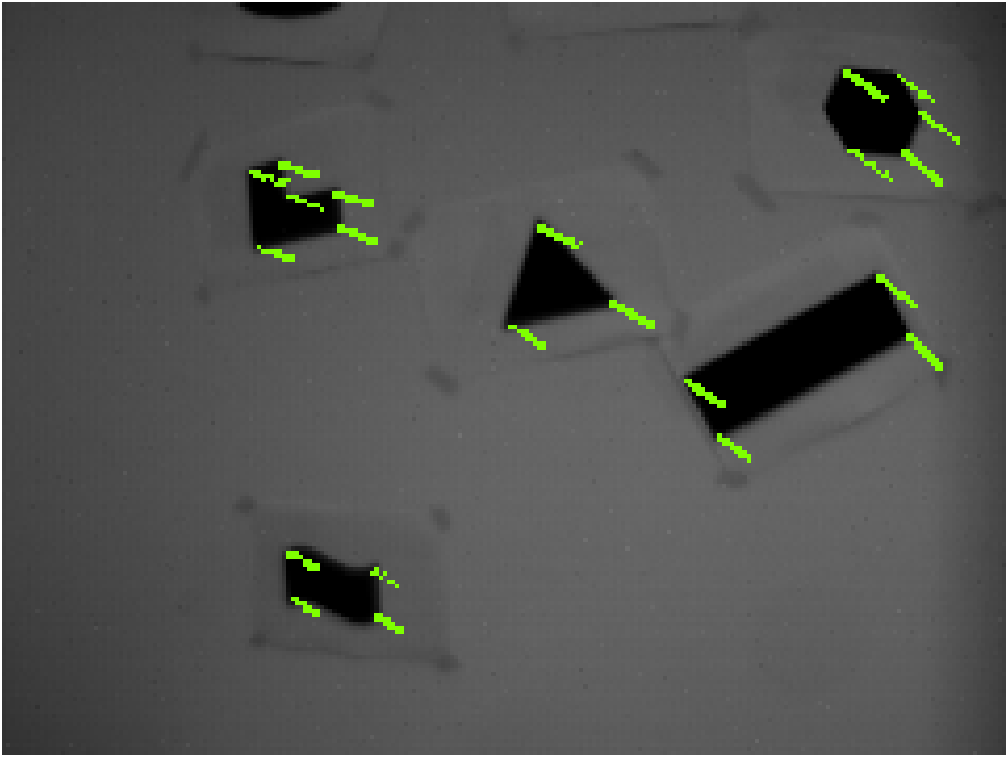}\label{gtshapes}}
    \caption{(a) The intensity corners detected by Harris with different parameters. (b) The ground-truth trajectories generated by the optimal parameters.}
    \label{steady_state}
\end{figure}

First of all, we propose to evaluate the ground-truth with the evHarris corner score. As an adaptation of Harris in events, evHarris is known to have the highest accuracy. Meanwhile, the Harris score is one of the most valid corner scores. With evHarris, we evaluate the number and score of ground-truth events. After that, we consider three parameters in Harris, which are the window size $w$ and standard deviation $\sigma$ of the Gaussian filter, and the detection quality $t_Q$.
Stronger smoothing (large $w$ and $\sigma$)  affects the location of the corners, whereas weaker smoothing causes the algorithm to be sensitive to noise. The $t_Q$ is defined as the ratio of the score threshold to the global maximum score. The higher threshold leads to more reliable corners, but also increases the miss detection rate. As shown in Fig. \ref{dist}, the intensity corners generated by different parameters are at different locations.

We prepare several regular candidates for each parameter and then perform a grid search. The candidates of $w$ are $\{3,5,7,9\}$, the candidates of $\sigma$ are $\{w/2,w/3,w/4\}$, and the candidates of $t_Q$ are $\{0.05,0.02,0.01,0.005\}$. We search for the parameter with the highest average score and the number of ground-truth events is not less than $80\%$ of the maximum number (all parameters in the scenario).

Similar to the previous works, only data between 100th to 400th frames are used. We label the events at the distance $[0,1]$ and $(1,5])$ from the intensity trajectory as positive and negative, respectively. By discarding events outside of $5$ pixels, we are able to cope with miss-detected intensity corners. This also ensures that sacrificing the number of ground-truth for quality hardly affects the evaluation. For the evaluation in Sec. \ref{sec:ExpEva}, we search for the optimal parameters for each of the four scenarios by grid search.

The optimal parameters for each scenario are reported in Table \ref{tab:gt}. As a comparison, we also report the ground-truth generated using Matlab's default parameters (with $w=5,\sigma=5/3,t_Q=0.01$). The ground-truth generated from the optimal parameters with \textit{shapes\_6dof} as an example is shown in Fig. \ref{gtshapes}, where the width of streams is $2$ pixels. We significantly improve the average score of the 2-pixels-width ground-truth while maintaining a certain quantity of it. More importantly, this evaluation method gives a overall view of ground-truth, and is applicable to other parameters and algorithms.

\begin{table}[t]
  \centering
  \caption{The number of ground-truth corner events detected per frame, and the average evHarris score of ground-truth events. The `*' stands for optimal parameters, `d' stands for the default parameters. \label{tab:gt}}
  \renewcommand{\arraystretch}{1.0}
    \begin{tabular}{l|m{1.4cm}<{\centering}|m{0.9cm}<{\centering}|m{0.9cm}<{\centering}|m{0.5cm}<{\centering}|m{0.5cm}<{\centering}}
    \hline
    scene& ($w^*$,$\sigma^*$,$t_Q^*$)  & $GT^*$ number & $GT_{d}$ number & $GT^*$ score     & $GT_{d}$ score \\ \hline
    \textit{shapes\_6dof}  & (3,$w$/4,0.05)  & 480.0        & 588.2   & \textbf{6.24} & 5.23      \\
    \textit{boxes\_6dof}   & (3,$w$/4,0.005) & 1739.7     & 1754.2  & \textbf{3.63} & 3.29      \\
    \textit{dynamic\_6dof} & (3,$w$/3,0.01)  & 976.1      & 1211.3  & \textbf{4.06} & 3.28      \\
    \textit{poster\_6dof}  & (5,$w$/4,0.01)  & 1804.1     & 2270.7  & \textbf{3.65} & 3.40      \\\hline
    \end{tabular}
  \end{table}

\begin{figure*}[t]
\centering
\includegraphics[width=1.0\textwidth]{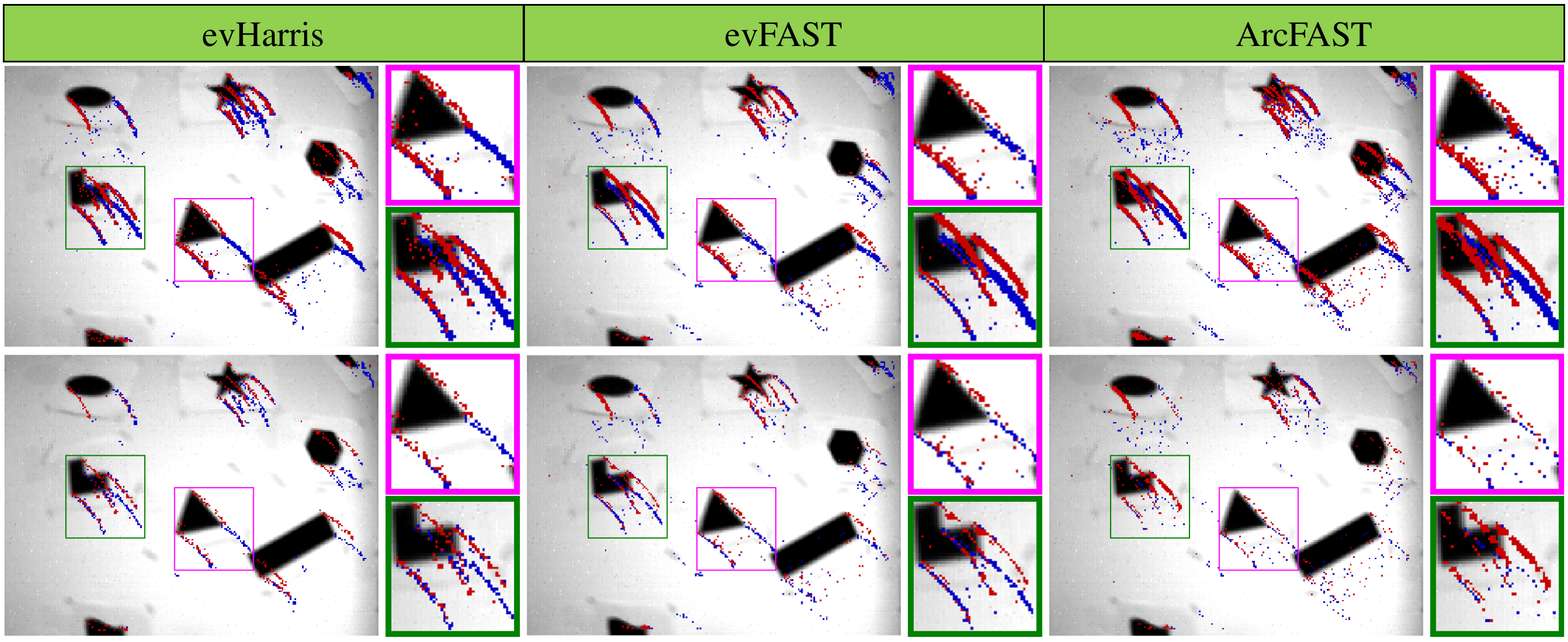}
\caption{The corner event stream detected by three detectors and their ANMS-based versions, with \textit{shapes-6dof} as an example. The top row is the raw detectors, and the bottom row is their ANMS-based versions. The ANMS suppresses coarse feature streams into fine corner streams.}
\label{fig::CORNERresult}
\end{figure*}

\section{Experiments and Evaluation}\label{sec:ExpEva}

We evaluated the proposed ANMS pipeline on the public DAVIS 240C dataset \cite{dataset}. This dataset is recorded with a DAVIS camera with a resolution of $240\times 180$, containing different camera movements in various scenes. To obtain the evaluation results of universal significance, we selected a representative subset according to the complexity of the scene (texture, lighting, dynamics, etc.), which include \textit{shapes\_6dof}, \textit{boxes\_6dof}, \textit{dynamic\_6dof}, \textit{poster\_6dof}. For the DAVIS 240C dataset, the motion becomes extremely intense in the later stages, which significantly reduces the effectiveness of the ground-truth. As in the previous works, the evaluation is performed between the 100th and 400th frames.

\begin{table}[t]
\centering
\caption{Reduction rates (\%) of different detectors and scenes. The `w' and `w/o' represent with and without ANMS respectively, and as blow. \label{tab:RR}}
  \begin{tabular}{l|cc|cc|cc}
    \hline
    \multirow{2}{*}{}      & \multicolumn{2}{c|}{evHarris}             & \multicolumn{2}{c|}{evFAST}                & \multicolumn{2}{c}{ArcFAST}               \\ \cline{2-7} 
                           & \multicolumn{1}{c|}{w/o}  & w             & \multicolumn{1}{c|}{w/o}   & w             & \multicolumn{1}{c|}{w/o}  & w             \\ \hline
    \textit{shapes\_6dof}  & \multicolumn{1}{c|}{8.07} & \textbf{1.42} & \multicolumn{1}{c|}{11.48} & \textbf{4.77} & \multicolumn{1}{c|}{8.00} & \textbf{2.85} \\ \hline
    \textit{boxes\_6dof}   & \multicolumn{1}{c|}{7.33} & \textbf{1.37} & \multicolumn{1}{c|}{3.51}  & \textbf{1.99} & \multicolumn{1}{c|}{6.23} & \textbf{3.00} \\ \hline
    \textit{dynamic\_6dof} & \multicolumn{1}{c|}{4.45} & \textbf{0.77} & \multicolumn{1}{c|}{2.93}  & \textbf{1.50} & \multicolumn{1}{c|}{5.70} & \textbf{2.72} \\ \hline
    \textit{poster\_6dof}  & \multicolumn{1}{c|}{6.89} & \textbf{1.35} & \multicolumn{1}{c|}{2.59}  & \textbf{1.53} & \multicolumn{1}{c|}{4.97} & \textbf{2.61} \\ \hline
    overall                & \multicolumn{1}{c|}{6.75} & \textbf{1.26} & \multicolumn{1}{c|}{3.72}  & \textbf{1.97} & \multicolumn{1}{c|}{5.89} & \textbf{2.82} \\ \hline
    \end{tabular}
\end{table}

\begin{figure}[t]
  \centering
  \subfloat[][]{\includegraphics[width=0.45\textwidth]{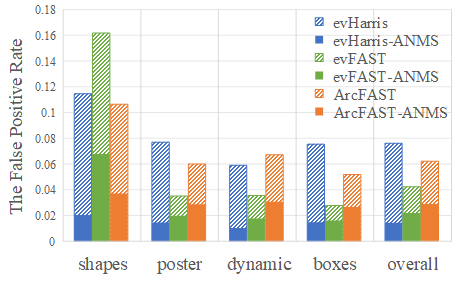}\label{fig::FPR1}}~\\
  \subfloat[][]{\includegraphics[width=0.4\textwidth]{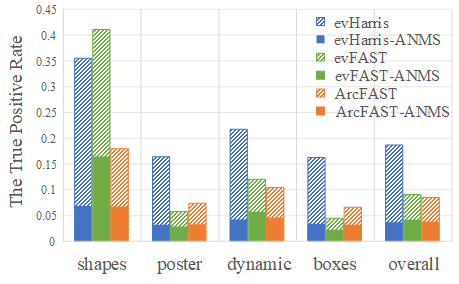}\label{fig::FPR2}}
  \caption{The false positive rate and true positive rate of each scene and detector.}
  \label{fig::FPR}
\end{figure}

\begin{table*}[t]
  \begin{center}
  \caption{Accuracy (\%) of different detectors on four datasets.\label{tab:ACC}}
  \begin{tabular}{l|cc|cc|cc|c}
    \hline
    \multirow{2}{*}{}      & \multicolumn{2}{c|}{evHarris}                       & \multicolumn{2}{c|}{evFAST}                         & \multicolumn{2}{c|}{ArcFAST}                     & \multicolumn{1}{c}{\multirow{2}{*}{FA-Harris}} \\ \cline{2-7}
                           & \multicolumn{1}{c|}{w/o}   & \multicolumn{1}{c|}{w} & \multicolumn{1}{c|}{w/o}   & \multicolumn{1}{c|}{w} & \multicolumn{1}{c}{w/o} & \multicolumn{1}{c|}{w} & \multicolumn{1}{c}{}                           \\ \hline
    \textit{shapes\_6dof}  & \multicolumn{1}{l|}{82.58} & \textbf{87.72}         & \multicolumn{1}{l|}{79.03} & \textbf{84.59}         & 81.34                   & \textbf{86.22}         & 86.48                                          \\ \hline
    \textit{boxes\_6dof}   & \multicolumn{1}{l|}{87.74} & \textbf{92.83}         & \multicolumn{1}{l|}{91.04} & \textbf{92.31}         & 88.80                   & \textbf{91.50}         & 93.00                                          \\ \hline
    \textit{dynamic\_6dof} & \multicolumn{1}{l|}{88.10} & \textbf{91.10}         & \multicolumn{1}{l|}{89.45} & \textbf{90.56}         & 86.41                   & \textbf{89.28}         & 90.88                                          \\ \hline
    \textit{poster\_6dof}  & \multicolumn{1}{l|}{87.29} & \textbf{92.07}         & \multicolumn{1}{l|}{90.95} & \textbf{91.88}         & 88.85                   & \textbf{90.96}         & 92.15                                          \\ \hline
    overall                & \multicolumn{1}{l|}{87.29} & \textbf{91.92}         & \multicolumn{1}{l|}{89.86} & \textbf{91.31}         & 87.85                   & \textbf{90.56}         & 91.90                                          \\ \hline
    \end{tabular}
    \captionsetup{justification=centering}
  \end{center}
\end{table*}

\begin{table}[t]
  \caption{The average time (ns) each event was processed by the three detectors and their ANMS-based versions.\label{tab:timeperf}}
    \centering
  \begin{tabular}{m{50pt}<{\centering}|m{30pt}<{\centering}|m{30pt}<{\centering}|m{30pt}<{\centering}|m{30pt}<{\centering}}
  \hline
  (ns/event) & w/o ANMS& w ANMS &Increase Time& Increase Rate\\\hline
  evHarris \cite{evHarris} & 13668 & 14282 & 613 & 4.48\%\\\hline
  evFAST \cite{evFAST}  & 3709  & 3853  & 144 & 3.88\%\\\hline
  ArcFAST \cite{ArcFAST} & 2145 & 2335   & 189 & 8.81\%\\\hline
  \end{tabular}
\end{table}

\subsection{Detection Performance}\label{sec:ktau}
We consider five detectors, namely evHarris \cite{evHarris}, evFAST \cite{evFAST}, ArcFAST \cite{ArcFAST}, FAHarris \cite{FAHarris}, and SITS \cite{SITS}. The FA-Harris itself is a coarse-to-fine detector, we compare our pipeline with it. Although the SITS detector has confidence scores that are applicable to our pipeline, the fine details of the algorithm are not available. Consequently, we evaluated ANMS in combination with evHarris, evFAST, and ArcFAST respectively. For evHarris and evFAST, we use the public implementation of Mueggler et al. \cite{evFAST}. For ArcFAST, we use the public implementation of Alzugaray et al. \cite{ArcFAST}. All of these implementations, including our method, are accomplished with C\texttt{++}.

First, we show the filtering result of ANMS on the three detectors using \textit{shapes\_6dof} as an example, as shown in Fig. \ref{fig::CORNERresult}. Overall, ANMS greatly reduces the width of the stream and outputs a finer stream. As mentioned earlier, the evFAST and ArcFAST use different algorithms to search for segments. When comparing their ANMS version, evFAST-ANMS outputs finer streams and continuous trajectories. This indicates that the designed scoring method is more effective for the evFAST algorithm. Our scoring method is not stable for the ArcFAST algorithm, which outputs frequently broken trajectories. The evHarris with Harris as the score also shows finer and continuous trajectories after ANMS. However, a certain amount of noise remains in ArcFAST, which will be discussed in Sec. \ref{limitations}.

Next, we use several criteria to evaluate the performance of these detectors, which are \textit{reduction-rate}, \textit{false-positive-rate}, \textit{true-positive-rate}, \textit{accuracy}. We report the results in Table \ref{tab:RR}, \ref{tab:ACC} and Fig. \ref{fig::FPR}.

First we compared the reduction rates of each detector in different scenarios and weighted them with the number of events to obtain the `overall' term. The reduction rate is defined as the proportion of the number of corner events to the total number of events. It indicates the ability of the detector to convert the event stream into the feature stream. Consistent with previous reports \cite{ArcFAST}, the overall reduction rate of evFAST is lower than that of ArcFAST, and the detectors have a higher reduction rate in the simple scenario. In principle, ANMS filters out non-corner events in the output of detectors. As shown in Table \ref{tab:RR}, ANMS does reduces the reduction rate for all three original detectors.

The ANMS does not interfere with the detection process of detectors and only filters the detection results. Ideally, ANMS removes all events that are not located in the center of the trajectory, which is verified to some extent in Fig. \ref{fig::CORNERresult}. We report the false positive rate and true positive rate for each detector in Fig. \ref{fig::FPR}. The results of each scene are weighted according to the event density to obtain the `overall' term. As shown in Fig. \ref{fig::FPR1}, ANMS significantly reduces the false positive rate for all detectors. However, this is followed by a decrease in the true positive rate, as shown in Fig. \ref{fig::FPR2}. The ArcFAST reported the same phenomenon in the experiment \cite{ArcFAST}, i.e., an isotropic change in the true positive rate and the false positive rate. One reason for this phenomenon is the bias of the ground-truth, which will be discussed in Sec. \ref{limitations}. Despite the imperfection of the ground-truth, we evaluated the effect of ANMS on the accuracy of detectors as suggested by FA-Harris. The accuracy is defined as the proportion of correctly classified samples. According to Table \ref{tab:ACC}, the ANMS improves the accuracy of all detectors to the same level as FA-Harris, especially in good contrast scenes. Note that, FA-Harris is only applicable to the evHarris, while our filter is universal to all asynchronous detectors.

\subsection{Computational Performance}
As a post-processing algorithm, the time performance of ANMS is equally essential. All detectors were run on the same computer, the experiments were performed on four datasets, and their average execution speed were reported. According to Table \ref{tab:timeperf}, the evHarris-ANMS spends $613ns$ more time to process each event on average. An important reason is that, the evHarris algorithm does not use conventional SAE. As a comparison, it only spends $144ns$ and $189ns$ more on evFAST and ArcFAST, both of them contain the SAE in detection. Nevertheless, the addition of ANMS only increased the time of the detectors by 4.49\%, 3.88\%, and 8.81\%, respectively. Therefore, it is concluded that as an event-based post-processing algorithm, ANMS has little effect on the time performance of original detectors.

\section{Discussion}\label{limitations}
In this section, we discuss the shortcomings of DAVIS-based ground-truth, the limitations of our approach, and directions for future work. First, the approach of using frame data to label event data has several limitations \cite{evFAST}. (i) As mentioned earlier, the frame data and event data have different interpretations for the corners. We mitigate this by evaluating the ground-truth and adjusting the parameters. (ii) The high dynamic range and high speed motion degrade the frame data quality. Therefore, only the data at the beginning were evaluated in the experiment. (iii) The intensity trajectory is limited by the frame rate. Alzugaray et al. \cite{ArcFAST} perform cubic splines interpolation on the intensity trajectory. However, the $2D$ trajectory cannot be accurately mapped to the $XYT$ space. We believe that improving DAVIS-based ground-truth requires fusing event clouds with intensity traces rather than using only frame data.

As shown in Fig. \ref{fig::CORNERresult}, our method cannot remove isolated noise of ArcFAST because the local DSSAE of isolated noise is small. Both NMS and ANMS aim to filter dense results with high response. Isolated noise is generally removed using spatial correlation-based filters instead of NMS. Moreover, even in evHarris and evFAST, ANMS still leads to some degree of discontinuity. A soft NMS is able to mitigates this problem by retaining the local maximum of $N$ events. Overall, in our opinion, further improvements on ground-truth and extensions of the ANMS pipeline to other asynchronous tasks are useful research directions for the event camera community.

\section{Conclusion}
The event-based non-maximum suppression is limited by the asynchrony of the event camera, which leads to the limitation of the event feature detector. The proposed pipeline performs NMS asynchronously by establishing an asynchronous corner response state DSSAE. In addition, ANMS is available for event-based FAST detectors with the proposed scoring methods for evFAST and ArcFAST. Similar to frame-based NMS, ANMS hardly affects the real-time performance of detectors. Most importantly, the proposed pipeline is generalizable to asynchronous scoring tasks in event vision. Finally, we improved the previous DAVIS-based labeling method by evaluating the ground-truth and tuning the parameters. The proposed method is available for additional parameter tuning to fill the gap between frames and events. We would like to improve the ground-truth and apply ANMS to more asynchronous tasks to take advantage of event cameras.

\bibliographystyle{IEEEtran}
\bibliography{ANMSRef.bib}

\end{document}